\documentclass[lettersize,journal]{IEEEtran}
\usepackage{amsmath,amsfonts}
\usepackage{amsmath}
\usepackage{algorithmic}
\usepackage{algorithm}
\usepackage{array}
\usepackage[caption=false,font=normalsize,labelfont=sf,textfont=sf]{subfig}
\usepackage{textcomp}
\usepackage{stfloats}
\usepackage{url}
\usepackage{verbatim}
\usepackage{graphicx}
\usepackage[hidelinks]{hyperref}
\usepackage{cite}
\usepackage{amssymb}
\usepackage{graphicx}
\graphicspath{{./figures/}} % 图片路径

\begin{document}

\title{Flexbee: A Grasping and Perching UAV Based on Soft Vector-Propulsion Nozzle}

\author{Yue Wang, Lixian Zhang,~\IEEEmembership{Fellow,~IEEE}, Yimin Zhu, Yangguang Liu, and Xuwei Yang
         % <-this % stops a space
\thanks{This work is supported in part by National Natural Science Foundation of China (623B2029). (Corresponding author:Lixian Zhang and Yimin Zhu.)}% <-this % stops a space
\thanks{Yue Wang, Lixian Zhang, Yimin Zhu, Yangguang Liu, and Xuwei Yang are with the School of Astronautics, Harbin Institute of Technology, Harbin 150001, China.(e-mail:  yuewang@stu.hit.edu.cn; lixianzhang@hit.edu.cn; ymzhu@hit.edu.cn; ygliu@stu.hit.edu.cn;25B904045@stu.hit.edu.cn).}}

% column for its text to clear the IEEEpubid mark.？？？？？？？

\maketitle

\begin{abstract}
The aim of this paper is to design a new type of grasping and perching unmanned aerial vehicle (UAV), Flexbee, characterized by its soft vector-propulsion nozzle (SVPN). Compared to previous UAVs, Flexbee integrates flight, grasping, and perching functionalities into the four SVPNs, offering advantages such as decoupled position and attitude control, high structural reuse, and strong adaptability for grasping and perching. A dynamics model of Flexbee has been developed, and the nonlinear coupling issue of the moment has been resolved through linearization of the equivalent moment model. Hierarchical control strategy was employed to design the controllers for Flexbee’s two operational modes. Finally, flight, grasping, and perching experiments were conducted to validate Flexbee’s kinematic capabilities and the effectiveness of the control strategy.
\end{abstract}%摘要

%关键词

\section{Introduction}
\IEEEPARstart{M}{ulti-rotor} unmanned aerial vehicles (UAVs), with their three-dimensional maneuverabilities, have demonstrated remarkable effectiveness in environments that are difficult for humans to reach \cite{ref1,ref2,ref3,ref4,ref5}. As people's requirements for UAV endurance performance and adaptability to complex environments offer greater advantages, compared with large UAVs, small UAVs have the characteristics of small size, light weight, low cost, and high maneuverability, which play a greater advantage in complex environments \cite{ref6,ref7,ref8}. However, conducting observation and surveillance work for a longer period of time with limited wind disturbance and battery capacity remains one of the main challenges in research. Inspired by bio-perching, perching has been proposed as a solution to save energy for flight \cite{ref9,ref10}.

Several perching mechanisms have been studied regarding perching: grasping perching, embedded perching, and adsorption perching \cite{ref11,ref12,ref13,ref14,ref15}. In particular, the mechanism for achieving the grasping perching is heavier than other methods, but is considered effective in natural environments \cite{ref16}. From the point of view of the method of the grasping perching, the grasping perching can be classified into passive grasping perching and active grasping perching.

In passive grasping perching, the grasping mechanism operates entirely passively, with the execution of the grasping action relying solely on the impact force of the aircraft to close the fixture, the mass of the UAV to maintain the closed state, and the lifting force of the aircraft exceeding gravity to open the fixture \cite{ref17,ref18,ref19}. Passive grasping eliminates the need for an actuator to perform the grasping action, thereby offering advantages such as low energy consumption and a simplified structural design in specific scenarios. However, in complex grasping scenarios, the shape and material of the graspable objects are often restricted by the grasping mechanism itself. Additionally, the allowable graspable angle is typically constrained to vertical grasping due to mechanical limitations, and attempting to perch at a large tilting angle in passive contact significantly reduces its success rate.   

In \cite{ref20,ref21,ref22,ref23,ref24,ref25}, active grasping mechanisms are employed, wherein motors or servos are utiliz\textbf{}ed to drive mechanical structures, enabling dynamic adjustment of the contact force and contact angle with the target. This allows adaptation to targets of varying shapes, materials, and positions. Although active grasping demonstrates superior adaptability compared to passive grasping, it typically involves additional independent actuators and more intricate mechanical structures to apply closed-loop control to the grasping action. Consequently, the increased system weight and complexity result in significantly higher overall energy consumption and reduced endurance for the UAVs.

In summary, passive grasping often struggles to adapt to complex grasping scenarios, while active grasping is possible but has low structural utilisation and high system complexity. In addition, both active and passive grasping technologies require the UAV to exhibit precise attitude control performance to ensure the feasibility of the grasping task. However, in previous research, the attitude coupling inherent in quadrotor-based grasping UAVs poses challenges in effectively demonstrating grasping performance. From the perspective of UAV structure, existing grasping UAVs typically feature entirely separate grasping and motion mechanisms, with the additional redundant mass leading to a low structural reuse rate. Meanwhile, the perch-grabbing mechanism mounted on a grasping and perching UAV is often significantly smaller relative to the UAV's overall volume, which restricts both the perch-grabbing range and the operational range of the UAV. A grasping and perching UAV that combines the high mobility, good structural reusability, and strong environmental adaptability remains an area of research that has yet to be explored.

\begin{figure}[!t]
\centering
\includegraphics[width=2.6in]{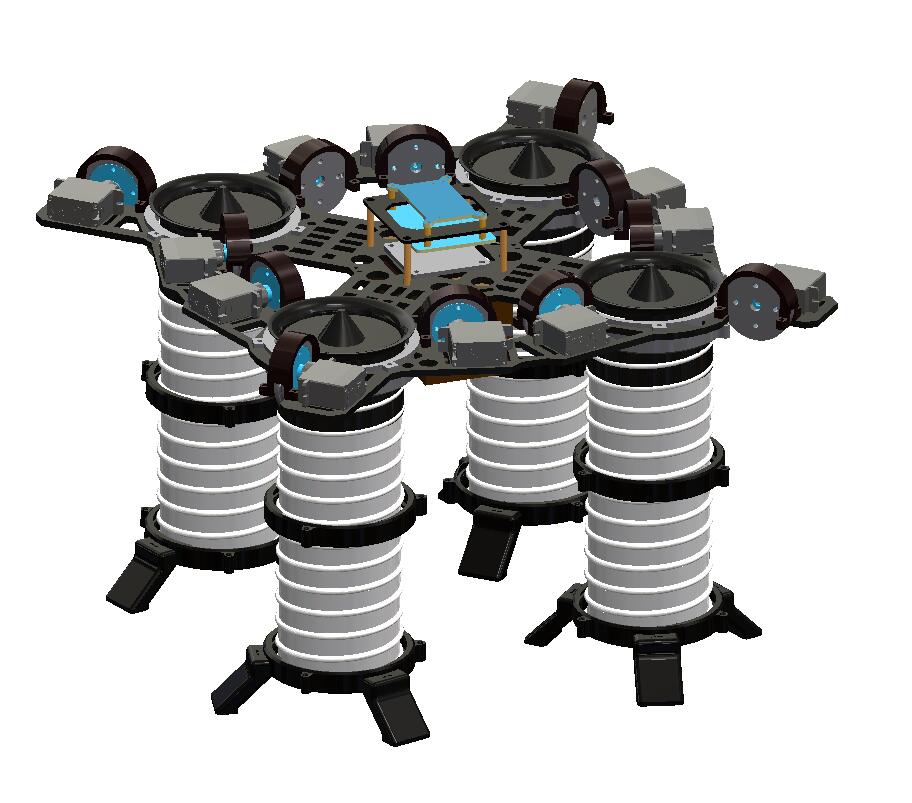}
\includegraphics[width=2.6in]{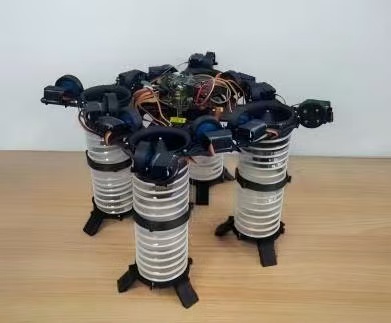}
\caption{Flexbee mechanical design and avionics}
\label{fig_1}
\end{figure}

Therefore, this study aims to fill this gap by designing a fully-actuated grasping and perching UAV based on the soft vector-propulsion nozzle (SVPN). The mechanical design, motion modeling, controller design, grasping platform motion strategy, and motion performance of this drone are comprehensively described. The mechanical design and electronic components of Flexbee are shown in Fig.\ref{fig_1}. The main contributions of this study are as follows:

\begin{enumerate}
\item We innovatively propose a grasping and perching UAV based on soft vector propulsion nozzles, named Flexbee. By manipulating the vectorial tilt of SVPNs and the differential speeds of the four rotors, Flexbee achieves decoupled position and attitude motion control in flight, while also enabling omnidirectional motion during the grasping/perching flight mode.

\item We propose the mechanical design of a soft vector propulsion nozzle along with its modeling and control methodology. SVPN enables control over the direction of the nozzle through wire-driven mechanism, generating vector thrust and moment. Simultaneously, it employs a curved hose for adaptive envelope grasping, achieving a breakthrough in integrating flight propulsion and grasping mechanisms, while significantly improving the structural utilisation rate.

\item We propose a linear fitting method for the nonlinear control allocation model based on equivalent torque, which significantly reduces the model complexity that makes it possible to decouple control allocation matrices that cannot be decoupled. Concurrently, we implement a hierarchical cooperative control strategy to manage both the individual components and the overall system in a hierarchical manner, and accordingly design controllers for the fully actuated flight mode and the grasping/perching flight mode.

\item Through motion experiments, grasping experiments, and perching experiments conducted on the physical platform, this study validates Flexbee's design scheme and control strategy.
\end{enumerate}

Flexbee presented in this study exhibits the following functions and advantages:

\begin{enumerate}
\item Grasping, perching, and motion functions are integrated within the soft-body structure, eliminating the need for additional mechanical structures beyond the motion actuators. The soft-body design enables Flexbee to perform multiple modal functions while significantly reducing its weight, and the synergistic design facilitates efficient energy reuse.

\item Flexbee is a fully-actuated UAV capable of tracking 6-D trajectories by adjusting the vector direction at the end of the soft body actuator, thereby decoupling position control and attitude control during flight. When performing grasping and perching functions, the UAV can adjust its position and attitude within a limited spatial range, enhancing its motion control capability and maneuverability to successfully complete the grasping and perching tasks.

\item The SVPN is capable of adjusting both the direction and curvature radius of the nozzle to achieve hand-like envelope grasping, which allows adaptation to objects of various sizes and perching environments. It can regulate the shape and volume of the object to be grasped while maintaining the aircraft's stability, and subsequently adaptively adjust the nozzle curvature to accomplish the grasping action once it enters the grasping range. After completing the grasping and perching maneuver, the UAV retains quadcopter-like controllability.
\end{enumerate}

The remaining sections of this study are organised as follows:Section \uppercase\expandafter{\romannumeral2} presents the mechanical structure design of the wire-driven soft-body vector nozzle and the grasping and  perching UAV. Section \uppercase\expandafter{\romannumeral3} introduces the kinematic model of the soft vector propulsion nozzle and the dynamics model of the grasping and perching UAV. Section  \uppercase\expandafter{\romannumeral4} details the controller design for the two operational modes of the grasping and  perching UAV. Section \uppercase\expandafter{\romannumeral5} includes the experimental results, comprising the grasping experiment, the perching experiment, and the flight experiment. Section \uppercase\expandafter{\romannumeral6} provides a conclusion to the paper.

\section{Mechanical design}
\begin{figure*}[!t]
\centering
\includegraphics[width=0.9\linewidth]{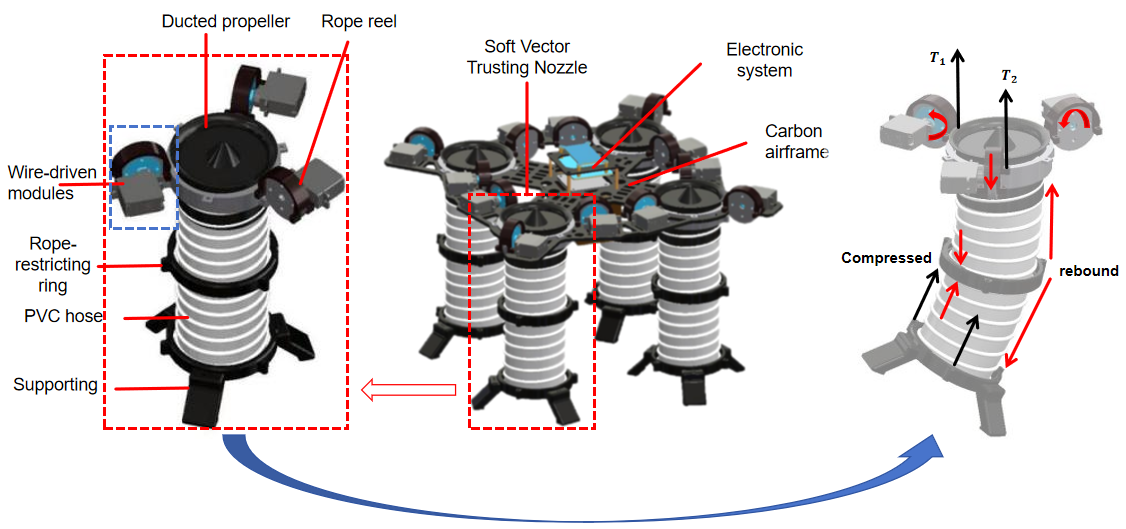}
\caption{Mechanical design of SVPN, mechanical design of Flexbee,and force analysis of SVPN.}
\label{fig_2}
\end{figure*}

This study will describe the mechanical designs of Flexbee and SVPN individually. The mechanical design of Flexbee is illustrated in Fig. \ref{fig_2}, comprising four SVPNs, a carbon airframe, and electronic systems. The arrangement of the four SVPNs references quadcopters, as depicted. The mechanical design of the SVPN is shown as Fig. \ref{fig_2} . It comprises one ducted propeller and three wire-driven modules as actuators, with PVC tubing and associated PLA components forming the primary structure. The SVPN utilizes a 12-bladed QF2611-4000KV ducted propeller for lift generation, employing DS031MG servo motors to adjust the rope length and thereby adjust the direction of the trailing end of SVPN.

For a SVPN, the three wire-driven modules are identical and evenly distributed along the grooves spaced around the circular carbon airframe. The initial length of the rope in each of the three wire-driven modules matches the original length of the hose. The rope is connected end-to-end with the SVPN and maintained in a pre-tensioned state.

\begin{figure*}[!t]
\centering
\includegraphics[width=0.9\linewidth]{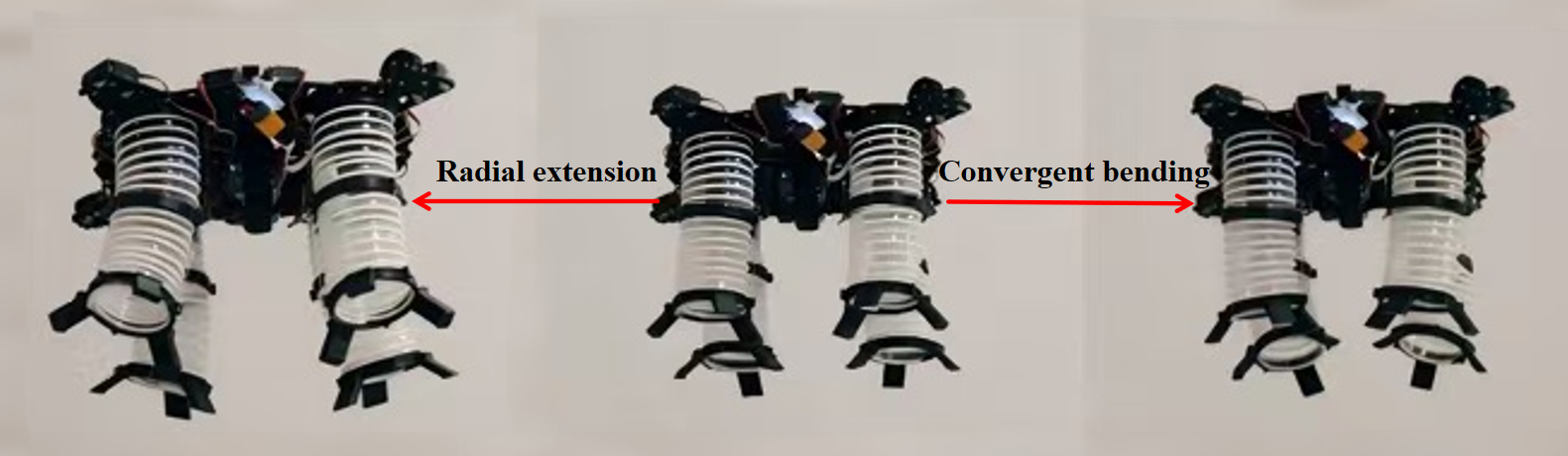}
\caption{Flexbee switches from fully-actuated flight mode to grasping/perching flight mode}
\label{fig_3}
\end{figure*}

Regarding the operating principle of the SVPN, as illustrated in Fig. \ref{fig_2}, when the Flexbee requires the generation of vector thrust, each SVPN must flex. The wire-driven modules actuate the corresponding side's cable to shorten it, rendering the length of the rope shorter than the original length of the SVPN on the same side. At this point, the taut rope exerts a tensile force perpendicular to the SVPN's cross-section. This force is transmitted to the SVPN, compelling it to undergo compressive deformation until its length equals that of the shortened rope. The other two ropes extend to match the deformed length, symmetrically distributing forces to counteract the torsional moment induced by the active rope's tension while constraining excessive deformation of the SVPN. At this equilibrium state, the tensile force in the rope equals and opposes the elastic force generated by the SVPN's deformation, thereby stabilising the SVPN's bent configuration. To ensure the terminal posture of the SVPN exhibits linear correlation with rope extension, kinematic modeling of the SVPN is required. Notably, when the bending direction lies at an angle between the ropes, simultaneous contraction of both ropes is necessary to generate tensile force. The bending direction and angle can be precisely adjusted by fine-tuning the lengths of the two ropes. Detailed modeling is presented in Section \uppercase\expandafter{\romannumeral3}.

In the mechanical design details, it is worth noting that a cable ring is threaded along the outer wall of the SVPN and rotated into position at the mid-section to ensure the end-point precision and structural strength of the SVPN control. At the same time, the servo motor is fitted with a rope reel and a guide cover to prevent the rope from slackening and detaching from the reel when the SVPN's endpoint is under load. These structural considerations aim to minimise discrepancies between the subsequent kinematic model and the physical prototype. The flexibility of the TPU material at each SVPN's end point is crucial for Flexbee taking off and landing, facilitating gas flow during ascent and providing cushioning during descent.

Flexbee achieves decoupled control of its attitude and position through the thrust and roll torque generated by the SVPN's vector thrust, as well as the pitch and yaw torques. The SVPN's rope-guided bending simultaneously creates gripping force and friction upon contact with grasped objects, enabling Flexbee to grasp and perch on tubular, plate-like and columnar objects. To adapt to the shapes of objects and perching environments, Flexbee pre-bends its nozzles outwards before grasping. This increases the grasping range, thereby enhancing adaptability and success rates. Notably, the Flexbee transitions from fully-actuated flight mode to grasping/perching flight mode during SVPN extension and bending. The grasping/perching mode controller maintains stable flight throughout this process.

In summary, the vector propulsion and flexible bending capabilities of the SVPN, combined with the coordinated operation of multiple SVPNs, give Flexbee fully-actuated flight and adaptive grasping capabilities.

\section{Dynamics modeling}
Before modeling the Flexbee, it is essential to establish the kinematic model of the SVPN. To ensure control accuracy, a linear mapping relationship must be defined between the position and attitude of the end of SVPN, and the length of the rope. At the same time, the SVPN's coordinate system must be aligned with that of the airframe because it involves determining the mapping relationship between the position and attitude of the end of SVPN and the airframe's coordinate system. The desired forces $F_{d}$ and moments $M_{d}$ generated by SVPN on the airframe can then be determined.
\subsection{Kinematic model of SVPN}
The following assumptions are made about the SVPN's actual physical properties before its kinematic modeling:

Assumption 1: The axis of the SVPN is a smooth, continuous curve with variable curvature over time.

Assumption 2: The SVPN produces almost no torsional deformation due to its structural limitations, so small torsional deformation is ignored.

Assumption 3: The SVPN does not produce significant radial size changes due to its structural limitations, so radial size deformation is ignored.

Assumption 4: A series of continuous cross-sections are formed when the end direction of the SVPN is changed. 

The kinematic model of the SVPN is consistent with the above assumptions and the piecewise constant curvature(PCC) modeling method. A detailed description can be found in the literature \cite{ref26,ref27}.

\begin{figure}
\centering
\includegraphics[width=0.7\linewidth]{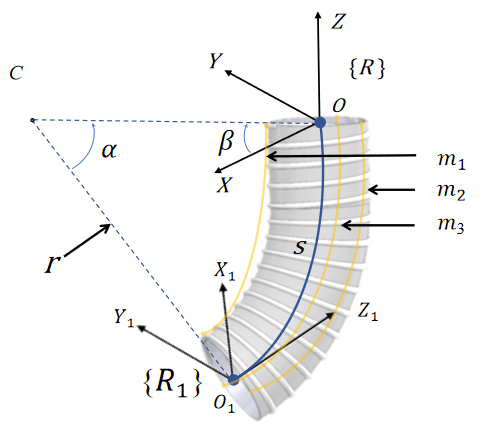}
\caption{Kinematic Description of SVPN.}
\label{fig_4}
\end{figure}

\begin{figure}
	\centering
	\includegraphics[width=0.7\linewidth]{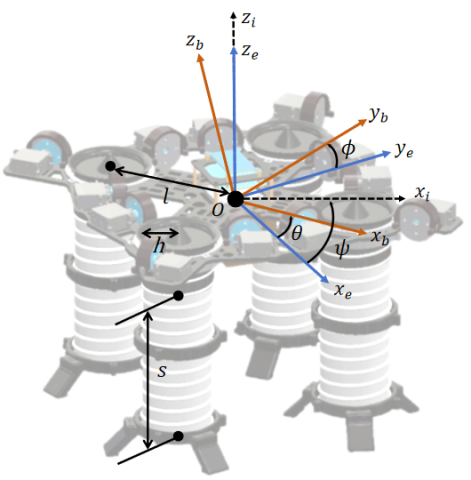}
	\caption{Definition of the Flexbee's coordinate system}
	\label{fig_5}
\end{figure}

In order to visually describe the kinematic model of SVPN, it is abstracted into the structure shown in Fig.\ref{fig_4}. The initial length of the rope is ${m_{ik}}$, ${i}$ denotes the number of SVPN, $k$ denotes the number of SVPN's rope, and $s_i$ is the initial length of the SVPN. SVPN is bent into a segment of arc in space, and $C$ denotes the centre of curvature of the arc. To describe the motion of the SVPN, a global coordinate system ${R}$ is introduced fixed at the centre of the propeller, and a local coordinate system ${R_1}$ is fixed at the centre of the end of SVPN, and the $Z$-axis of ${R}$ and ${R_1}$ are perpendicular to the respective circular planes, as shown in the Fig.\ref{fig_5}. According to the assumption of segmented constant curvature modeling the state of SVPN moving in space can be fully defined by a set of curvilinear parameters $\kappa_i= [r_i \ \alpha_i \ \beta_i ]^T $, with a radius of curvature $ r_i \in (0, +\infty) $, a centre of curvature $C$, an angle of bending $\alpha_i$, and an angle of inclination of the plane with respect to the positive direction of the $X$-axis $ \beta_i $.

To obtain the relationship between the position of the end of SVPN and the amount of change in the length of the rope, three spaces and two mappings are defined. Among them, the drive space is denoted as $[m_{i1} \ m_{i_2} \ m_{i_3} ]^T$,the configuration space is denoted as $ [r_i \ \alpha_i \ \beta_i ]^T $, and the task space $ [x_i \ y_i \ z_i ]^T$. The mapping from the drive space to the configuration space is established as %\eqref{eq:a}
%\eqref{eq:r} \eqref{eq:beta}
% Requires: \usepackage{amsmath}
\begin{equation}
    \alpha_i = \frac{2\sqrt{m_{i1}^2 + m_{i2}^2 + m_{i3}^2 - m_{i1}m_{i2} - m_{i1}m_{i3} - m_{i2}m_{i3}}}{3h},
    \label{eq:a}
\end{equation}
\begin{equation}
    r_i = \frac{(m_{i1} + m_{i2} + m_{i3})h}{2\sqrt{m_{i1}^2 + m_{i2}^2 + m_{i3}^2 - m_{i1}m_{i2} - m_{i1}m_{i3} - m_{i2}m_{i3}}},
    \label{eq:r}
\end{equation}
\begin{equation}
    \beta_i = 
    \begin{cases} 
      \arctan\left(\frac{m_{i2} + m_{i3} - 2m_{i1}}{\sqrt{3(m_{i2} - m_{i3})}}\right) & \text{if } x_{\text{end}} \geq 0, \, y_{\text{end}} \geq 0 \\
      \arctan\left(\frac{m_{i2} + m_{i3} - 2m_{i1}}{\sqrt{3(m_{i2} - m_{i3})}}\right) + \pi & \text{if }  x_{\text{end}}< 0 \\
      \arctan\left(\frac{m_{i2} + m_{i3} - 2m_{i1}}{\sqrt{3(m_{i2} - m_{i3})}}\right) + 2\pi & \text{if } x_{\text{end}} > 0, \ y_{\text{end}} < 0
    \end{cases}
    \label{eq:beta}
\end{equation}
where $x_{\text{end}}$ and $y_{\text{end}}$ are the position of the end of SVPN. 
 
Consider the mapping relationship from the configuration space to the task space, i.e., solving the sub-transformation matrix $H$ from ${R_1}$ to ${R}$. $H$ can be obtained by the following five steps. First, rotate $\beta$ around the $Z$-axis of ${R}$; then move $r$ along the positive $X$-axis under the current reference; then rotate $\alpha$ around the $Y$-axis of the current reference system; then move $r$ along the negative $X$-axis of the current reference system; and finally rotate $-\beta$ around the $Z$-axis of the current coordinate system, and finally obtain the homogeneous transformation matrix from the configuration space to the task space as 
% Requires: \usepackage{amsmath}
\begin{equation}
    H(\alpha_i, \beta_i, r_i) = \begin{bmatrix} 
    \mathbf{Q}{(\kappa)} & \mathbf{P}{(\kappa)} \\ 
    0 & 1 
    \end{bmatrix}
    \label{eq:matrixH}
\end{equation}
where $Q(\alpha_i,\beta_i) \in R_{3\times3} $ is denoted as the rotation matrix from ${R_1}$ to ${R}$, denoted as 
% Requires: \usepackage{amsmath}
\begin{equation}
    Q(\alpha_i,\beta_i) = \begin{bmatrix}
         C^2_{\beta}C_{\alpha}+S^2_{\beta} &  C_{\beta}S_{\beta}C_{\alpha} - C_{\beta}  S_{\alpha} & C_{\beta} S_{\alpha} \\
         C_{\beta}  S_{\beta}  C_{\alpha} -  C_{\beta} S_{\alpha} &  S^2_{\beta}  C_{\alpha} +  C^2_{\beta} &  S_{\beta} S_{\alpha} \\
        - C_{\beta} S_{\alpha} & - S_{\beta}  S_{\alpha} &  C_{\alpha}
    \end{bmatrix}.
\label{eq:rotation_matrix}
\end{equation}

$P(\kappa) \in R_{3\times1}$ denotes the positional translation vector from ${R_1}$ to ${R}$ , given by 
% Requires: \usepackage{amsmath}
\begin{equation}
    \mathbf{P}(\kappa) = 
    \begin{bmatrix}
        r_i \cos \beta_i (1 - \cos \alpha_i) \\
        r_i \sin \beta_i (1 - \cos \alpha_i) \\
        r_i \sin \alpha_i
    \end{bmatrix}.
    \label{eq:placeholder_label}
\end{equation}

It follows that the direction of the end of SVPN can be represented by the projection $Q(\alpha_i,\beta_i)$ in the ${R}$, and the centre of the circle at the end of SVPN can be represented by the representation, i.e., the positive kinematics model of SVPN.
According to the geometrical relations of SVPN shown i n Fig. \ref{fig_4}, the lengths of all ropes $[m_{i1} \ m_{i2} \ m_{i3} ]^T$ can be calculated as 
% 正文区代码
\begin{align}
m_{i1} &= \alpha_i \cdot \left( r_i - h \sin\beta_i \right) \label{eq:m1}, \\
m_{i2} &= \alpha_i \cdot \left[ r_i - h \sin\left( \beta_i - \frac{2\pi}{3} \right) \right], \label{eq:m2} \\
m_{i3} &= \alpha_i \cdot \left[ r_i - h \sin\left( \beta_i + \frac{2\pi}{3} \right) \right]. \label{eq:m3}
\end{align}

Due to their different positions in the airframe, the mapping relationship of each SVPN is different. The airframe can be rotated by pairs to derive the length equations of different SVPNs, i.e., $\beta_i = \beta_{i-1}+ {(i-1)}90^{\circ}$.
The ducted propeller is fixed to carbon airframe and drives the paddles to provide $F_{bi}$, let $\omega_i$ denote the $i$th ducted propeller's rotational speed, it generates $Z$-axis positively oriented force $T_i= c_t\omega_i^2$, the vector thrust$[F_{xi} \ 
 F_{yi} \ F_{zi}]^T$ is expressed as 
\begin{equation}
\begin{bmatrix} F_{x_i} \\ F_{y_i} \\ F_{z_i} \end{bmatrix} = Q(\kappa) \begin{bmatrix} 0 \\ 0 \\ T_i \end{bmatrix} = \begin{bmatrix} T_i \cos\beta_i \sin\alpha_i \\ T_i \sin\beta_i \sin\alpha_i \\ T_i \cos\alpha_i \end{bmatrix} = \boldsymbol{B} \begin{bmatrix} \alpha_i \\ \beta_i \\ \omega_i \label{eq:B} \end{bmatrix}. \end{equation}

There is a unique mapping relationship between $[\alpha_i \ \beta_i \ \omega_i]$ and $[F_{xi} \ 
 F_{yi} \ F_{zi}]$ where $B \in R_{3\times3}$. 
$d_i$ denotes the position vector of the $i$th SVPN with respect to the airframe, and the moment can be expressed as 
\begin{equation}
\boldsymbol{M_i} = \boldsymbol{F_i} \times \boldsymbol{d_i}.
\label{eq:torque}  % 用于交叉引用
\end{equation}

The collation from \eqref{eq:m1} \eqref{eq:m2} \eqref{eq:m3} \eqref{eq:B} \eqref{eq:torque} gives the mapping relation between $m$ and $\boldsymbol{F}$, $\boldsymbol{M}$.

 \subsection{Dynamics Model of Flexbee}
The dynamics model of Flexbee consists of a fully-actuated flight mode and an under-actuated grasping/perching flight mode which can be shown in
\begin{equation}
\begin{bmatrix} M & 0_3 \\ 0_3 & I \end{bmatrix}
\begin{bmatrix} \dot{\boldsymbol{r}} \\ \dot{\boldsymbol{\omega}} \end{bmatrix}
 = 
\begin{bmatrix} -Mg \\ -\boldsymbol{\omega} \times I\boldsymbol{\omega} \end{bmatrix}
 + 
\begin{bmatrix} R & 0_3 \\ 0_3 & I_3 \end{bmatrix}
\begin{bmatrix} \boldsymbol{F} \\ \boldsymbol{\tau} \end{bmatrix},
\label{eq:matrix_equation}
\end{equation}
where $M$, $I$, $r$, $\omega$, $g$, $R$ and $m$ denote the mass matrix, inertia matrix, moment of inertia of the aircraft around its centre of mass, angular velocity, gravity vector, rotation matrix and mass respectively. In addition, $F$ and $\tau$ denote the force and moment acting on the centre of mass, respectively.

Considering the symmetry of the quadrotor and the SVPN, we have the following dynamics equations, eq. \eqref{eq:xyz}, where the translational dynamics are represented in the inertial system and the attitude dynamics are represented in the machine system.
\begin{equation}
\begin{aligned}
m\ddot{x} &= C_{\theta}S_{\psi}F_x + (S_{\psi}S_{\theta}S_{\phi} + C_{\phi}C_{\psi})F_y + \\&\quad (S_{\psi}S_{\theta}C_{\phi} - S_{\phi}C_{\psi})F_z, \\
m\ddot{y} &= C_{\theta}C_{\psi}F_x + (C_{\psi}S_{\theta}S_{\phi} - C_{\phi}S_{\psi})F_y + \\&\quad (C_{\psi}S_{\theta}C_{\phi} + S_{\phi}S_{\psi})F_z, \\
m\ddot{z} &= mg + S_{\theta}F_x - S_{\phi}C_{\theta}F_y - C_{\phi}C_{\theta}F_z, \\
\ddot{\phi} &= \frac{1}{I_{xx}}\left[ M_x + qr(I_{yy} - I_{zz}) \right], \\
\ddot{\theta} &= \frac{1}{I_{yy}}\left[ M_y + pr(I_{zz} - I_{xx}) \right], \\
\ddot{\psi} &= \frac{1}{I_{zz}}\left[ M_z + pq(I_{xx} - I_{yy}) \right],
\end{aligned}
\label{eq:xyz}
\end{equation}
where $\phi$, $\theta$, and $\psi$ denote the roll, pitch, and yaw angles, respectively, and $p$, $q$ and $r$ denote the angular velocity of the airframe along the $X$, $Y$ and $Z$ axes, respectively. In addition $S_{\phi}$, $C_{\phi}$, $S_{\theta}$, $C_{\theta}$, $S_{\psi}$, $C_{\psi}$ is the abbreviated form of sine and cosine functions, and the position of the airframe in the inertial system is denoted by $x$, $y$, and $z$. $F_x$, $F_y$, $F_z$ are the forces acting on the $X$, $Y$, and $Z$ axes of the airframe and are all synthesised from the $F_{xi}$, $F_{yi}$,and $F_{zi}$ produced by the individual SVPNs, as shown in
\begin{equation}
\begin{aligned}
F_{bx} &= F_{bx1} + F_{bx2} + F_{bx3} + F_{bx4}, \\
F_{by} &= F_{by1} + F_{by2} + F_{by3} + F_{by4}, \\
F_{bz} &= F_{bz1} + F_{bz2} + F_{bz3} + F_{bz4}.
\end{aligned}
\label{eq:F}
\end{equation}
$M_{xi}$, $M_{yi}$, $M_{zi}$ are the force distances acting on the $X$, $Y$, and $Z$ axes of the airframe, and are all derived by cross-multiplying the $F_{xi}$, $F_{yi}$, $F_{zi}$ produced by the individual SVPNs with the $d_i$.

\begin{figure}
\centering
\includegraphics[width=0.83\linewidth]{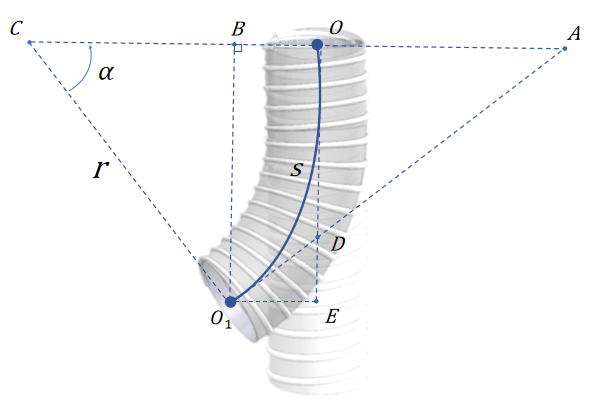}
\caption{Schematic diagram of a linear approximation to a non-linear model}
\label{fig_6}
\end{figure}

The vector force generated by each SVPN on Flexbee is only related to the nozzle attitude, but the position of SVPN affects the lever arm of the thrust vector acting on the aircraft, resulting in a strongly coupled torque vector with nonlinear characteristics, making decoupling difficult and independent control of attitude impossible. 

Therefore, a moment equivalence method is used to perform a linear approximation of the nonlinear model, employing an equivalent moment action point to eliminate the influence of position on the moment. The following is the specific modeling and decoupling process. As shown in Figure \ref{fig_6}, $O$ is the origin of the coordinate system, $O_1$ is the original point of action of the vector thrust, $C$ is the centre of curvature, $A$ is the centre of mass of the airframe, and $D$ is the intersection point of $O_1A$ and $OE$. When the shape of SVPN changes, $D$ moves along $OE$, and the vector thrust remains the same as the moment generated by $O_1$ and $D$ on the centre of the airframe. Let $OD = a$. Based on the similarity of triangles, we can derive that $a = s \cdot \frac{\tan\left(\frac{\alpha}{2}\right)}{\alpha}$. When $0^+$, its Taylor series expansion yields $a_{min} = 0.5$. Within the range of $(0,45^\circ)$, using the derivative properties for analysis, we obtain $a_{max} = 0.527s$. Therefore, $a = 0.5135s$ is taken as the equivalent vector thrust action point, with an error range of less than $2.7\%$. This moment equivalence approximation method allows the moment action point of the vector thrust to remain fixed at $D$, while changes at the end of SVPN only alter the direction without changing the position.$M_{xi}$, $M_{yi}$, $M_{zi}$ is shown as 
\begin{equation}
\begin{aligned}
M_x &= -0.5135s(F_{by1} + F_{by2} + F_{by3} + F_{by4}) \\
    &\quad - \left(\frac{\sqrt{2}}{2}l\right)(F_{bz1} + F_{bz2} - F_{bz3} - F_{bz4}), \\[1ex]  % 增加行间距
M_y &= 0.5135s(F_{bx1} + F_{bx2} + F_{bx3} + F_{bx4}) \\
    &\quad + \left(\frac{\sqrt{2}}{2}l\right)(F_{bz1} - F_{bz2} - F_{bz3} + F_{bz4}), \\[1ex]
M_z &= \left(\frac{\sqrt{2}}{2}l\right)\left[(F_{bx1} - F_{by1}) + (F_{bx2} + F_{by2}) \right. \\
    &\quad \left. - (F_{bx3} - F_{by3}) - (F_{bx4} + F_{by4})\right],
\end{aligned}
\label{eq:M}
\end{equation}
where $l$ is the distance between the SVPN and the centre of the airframe and $s$ is the axial length of the SVPN. Since Flexbee only needs to do kinematic attitude control of the end of SVPN, $s$ can be taken to be a constant value.

Therefore, the moments $M_{xi}$, $M_{yi}$, $M_{zi}$ can be uniquely mapped by $F_{xi}$, $F_{yi}$, $F_{zi}$ i.e., both forces and moments of the airframe can be mapped by the vector thrusts of the four SVPNs. Combining \eqref{eq:F} \eqref{eq:M}, the complete dynamics model can be obtained.

\section{Multimodal control}
This section presents the controller architecture for Flexbee's two operational modes: fully-actuated flight mode and under-actuated grasping/perching flight mode. It is worth noting that these modes can be switched during flight, with the aircraft's attitude and position stability being maintained throughout. The two controllers are detailed separately in the following subsections.

\subsection{Fully-actuated flight mode controller}

\begin{figure}
\centering
\includegraphics[width=1\linewidth]{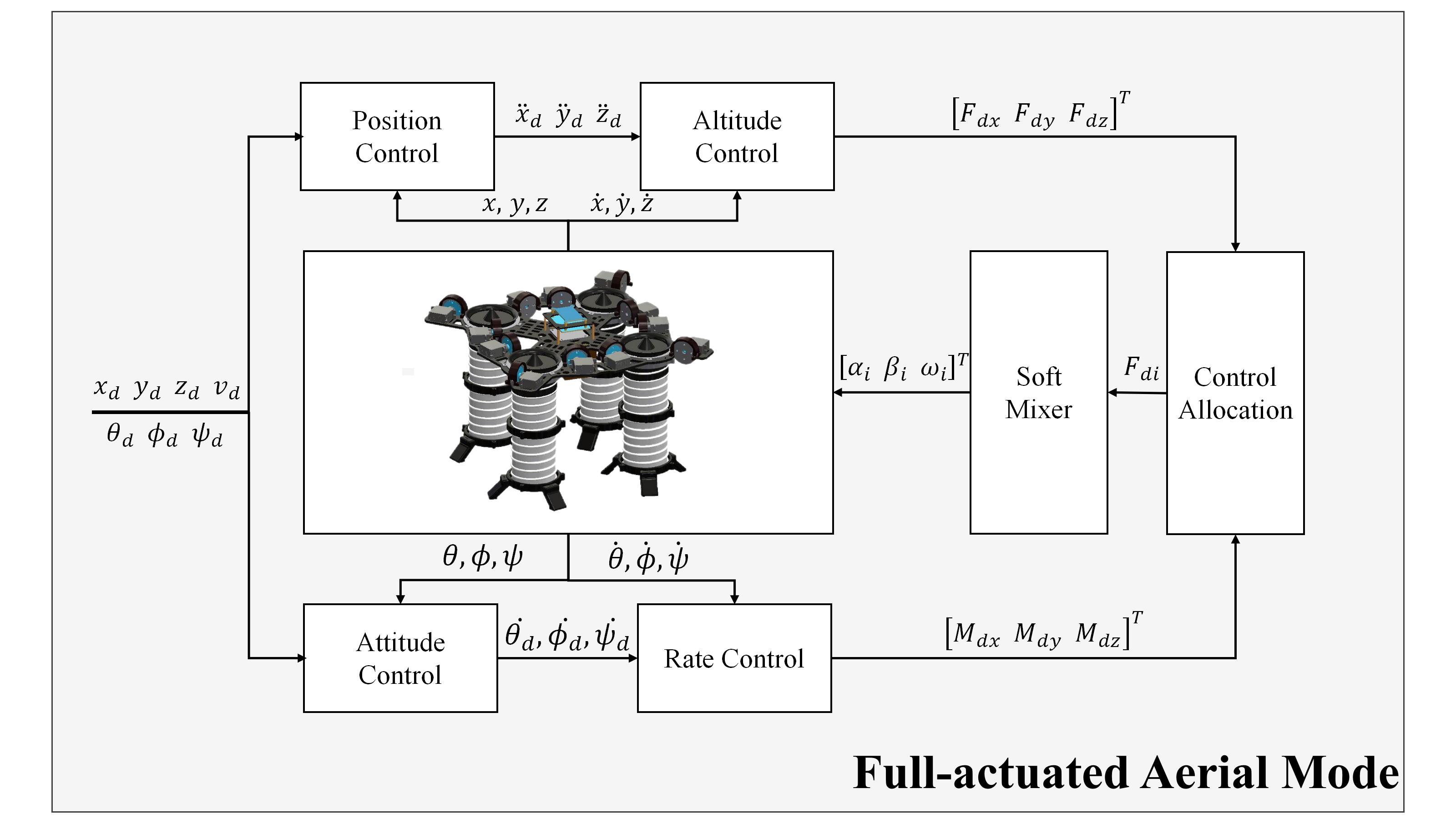}
\caption{The architecture of Fully-actuated flight mode controller}
\label{fig_7}
\end{figure}
Details of several core modules in the fully-actuated flight mode controller as shown in the figure are as Fig.~\ref{fig_7}.

\textbf{Position, Attitude, Velocity, Angular Velocity Controller:} The fully-actuated flight mode controller receives the desired position and desired velocity in the X, Y, and Z axes, as well as the desired attitude angle from the advanced planner. In the position, attitude, velocity, and angular velocity control loops, we use the following Proportional Integral Derivative (PID) laws:
\begin{equation}
u = K_{p,r}^a (\tau_d - \tau) + K_{d,r}^a (\dot{\tau}_d - \dot{\tau}) + K_{i,r}^a \int (\tau_d - \tau) d\tau,
\label{eq:pid}
\end{equation}
where $\tau = \{x, y, z, \dot{x}, \dot{y}, \dot{z}, \phi, \theta, \psi, \dot{\phi}, \dot{\theta}, \dot{\psi}\}$ and $u$ denotes the corresponding output for different $\tau$ .

\textbf{Control Allocator:} The desired force ${[F_{dx} \ F_{dy} \ F_{dz}]}^T$  and desired moment ${[M_{dx} \ M_{dy} \ M_{dz}]}^T $ are output to the control allocator by the velocity controller and angular velocity controller, respectively. The control allocation matrix in the control allocator is given by \eqref{eq:aF}:  

\begin{equation}
\left[ 
\begin{matrix} F_x \\ F_y \\ F_z \\ M_x \\ M_y \\ M_z \end{matrix}
\right] 
= 
A 
\left[ 
\begin{matrix} 
\begin{matrix} F_{bx1} \\ F_{by1} \\ F_{bz1} \end{matrix} \\ 
\begin{matrix} F_{bx2} \\ F_{by2} \\ F_{bz2} \end{matrix} \\ 
\begin{matrix} F_{bx3} \\ F_{by3} \\ F_{bz3} \end{matrix} \\ 
\begin{matrix} F_{bx4} \\ F_{by4} \\ F_{bz4} \end{matrix}
\end{matrix} 
\right],
	\left[ 
	\begin{matrix} 
		\begin{matrix} F_{bx1} \\ F_{by1} \\ F_{bz1} \end{matrix}
		\\ 
		\begin{matrix} F_{bx2} \\ F_{by2} \\ F_{bz2} \end{matrix}
		\\ 
		\begin{matrix} F_{bx3} \\ F_{by3} \\ F_{bz3} \end{matrix}
		\\ 
		\begin{matrix} F_{bx4} \\ F_{by4} \\ F_{bz4} \end{matrix}		
	\end{matrix} 
	\right]= A^+\left[ 
	\begin{matrix} F_x \\ F_y \\ F_z \\ M_x \\ M_y \\ M_z \end{matrix}
	\right]
	\label{eq:aF}.
\end{equation}

The control allocator uses a hierarchical control strategy in which the triaxial forces and moments acting on the airframe are mapped onto the  forces generated by each nozzle via matrix A, as shown in \eqref{eq:A}. The parameter expressions for rows 4–6 depend entirely on the airframe's structural parameters, meaning that A is a constant-parameter matrix.

\begin{figure*}[ht]
    \centering
    \begin{equation}
        \boldsymbol{A} = 
        \left[\begin{array}{cccccccccccc}  % 12列，居中对齐
            1      & 0        & 0        & 1      & 0        & 0        & 1      & 0        & 0        & 1      & 0        & 0        \\
            0      & 1        & 0        & 0      & 1        & 0        & 0      & 1        & 0        & 0      & 1        & 0        \\
            0      & 0        & 1        & 0      & 0        & 1        & 0      & 0        & 1        & 0      & 0        & 1        \\
            0      & -a   & -\dfrac{\sqrt{2}}{2}l & 0      & -a   & -\dfrac{\sqrt{2}}{2}l & 0      & -a   & \dfrac{\sqrt{2}}{2}l & 0      & -a   & \dfrac{\sqrt{2}}{2}l \\
            a  & 0        & \dfrac{\sqrt{2}}{2}l & a  & 0        & -\dfrac{\sqrt{2}}{2}l & a  & 0        & -\dfrac{\sqrt{2}}{2}l & a  & 0        & \dfrac{\sqrt{2}}{2}l \\
            \dfrac{\sqrt{2}}{2}l & -\dfrac{\sqrt{2}}{2}l & 0      & \dfrac{\sqrt{2}}{2}l & \dfrac{\sqrt{2}}{2}l & 0      & -\dfrac{\sqrt{2}}{2}l & \dfrac{\sqrt{2}}{2}l & 0      & -\dfrac{\sqrt{2}}{2}l & -\dfrac{\sqrt{2}}{2}l & 0        \\
        \end{array}
\right] 
\label{eq:A} % 正确闭合array环境
    \end{equation} % 可选标签（用于引用）
    \hrule % 添加横线
\end{figure*}

Since the rows and columns of A are not equal and only the rows are full rank, rank(A) = 6 .It is a non-square matrix, so it is necessary to find the pseudo-inverse matrix of it $A^+=A^T(AA^T)^{-1}$ . 

Then the forces of each SVPN can be expressed by the desired force and moment acting on the airframe as \eqref{eq:aF}.

The desired force $F_{bi}$ of each SVPN acting on the airframe can thus be obtained by the control allocator by controlling the distribution of $F_{d}$ and $M_{d}$ received by the body and passing them to the next control layer.

\textbf{SVPN soft mixer:} The control mixer for the second layer of hierarchical control, from \eqref{eq:alpha}\eqref{eq:betab}\eqref{eq:omega}, $\begin{bmatrix} \alpha_i & \beta_i & \omega_i \end{bmatrix}^T$ and $\begin{bmatrix} F_{xi} & F_{yi} & F_{zi} \end{bmatrix}^T$ unique mapping relationship, the matrix B is orthogonal matrix, so it can be obtained by its inverse,that is
\begin{align}
  &\alpha_i = \arctan\left( \frac{\sqrt{F_{xi}^2 + F_{yi}^2}}{F_{zi}} \right), \label{eq:alpha}\\
  &\beta_i = \arctan\left( \frac{F_{yi}}{F_{xi}} \right), 
  \label{eq:betab}\\
  &\omega_i = \sqrt[4]{\frac{F_{xi}^2 + F_{yi}^2 + F_{zi}^2}{c_\tau^2}}.
  \label{eq:omega}
\end{align}

The desired angle and rotational speed of each unit can be obtained from Eq. and then act on the body to produce the actual force and moment.
Through this control allocation strategy of hierarchical cooperative control, the rotational speed $\omega$ of the ducted propeller and the nozzle inclination angles $\alpha$ and $\beta $ can be calculated based on the desired triaxial forces and moments, and then the mapping relationship between the triaxial forces and moments and $m$ can be obtained from Eq. a. This means that the complete control allocation framework is obtained.

\subsection{Under-actuated grasping/perching flight controller}

Unlike the fully-actuated flight mode controller, the under-actuated grasping/perching flight controller adds an additional flight control model when Flexbee is in current mode. At the same time, it switches the controller frame and parameters to under-actuated flight mode. This is detailed below.

\begin{figure}
\centering
\includegraphics[width=1\linewidth]{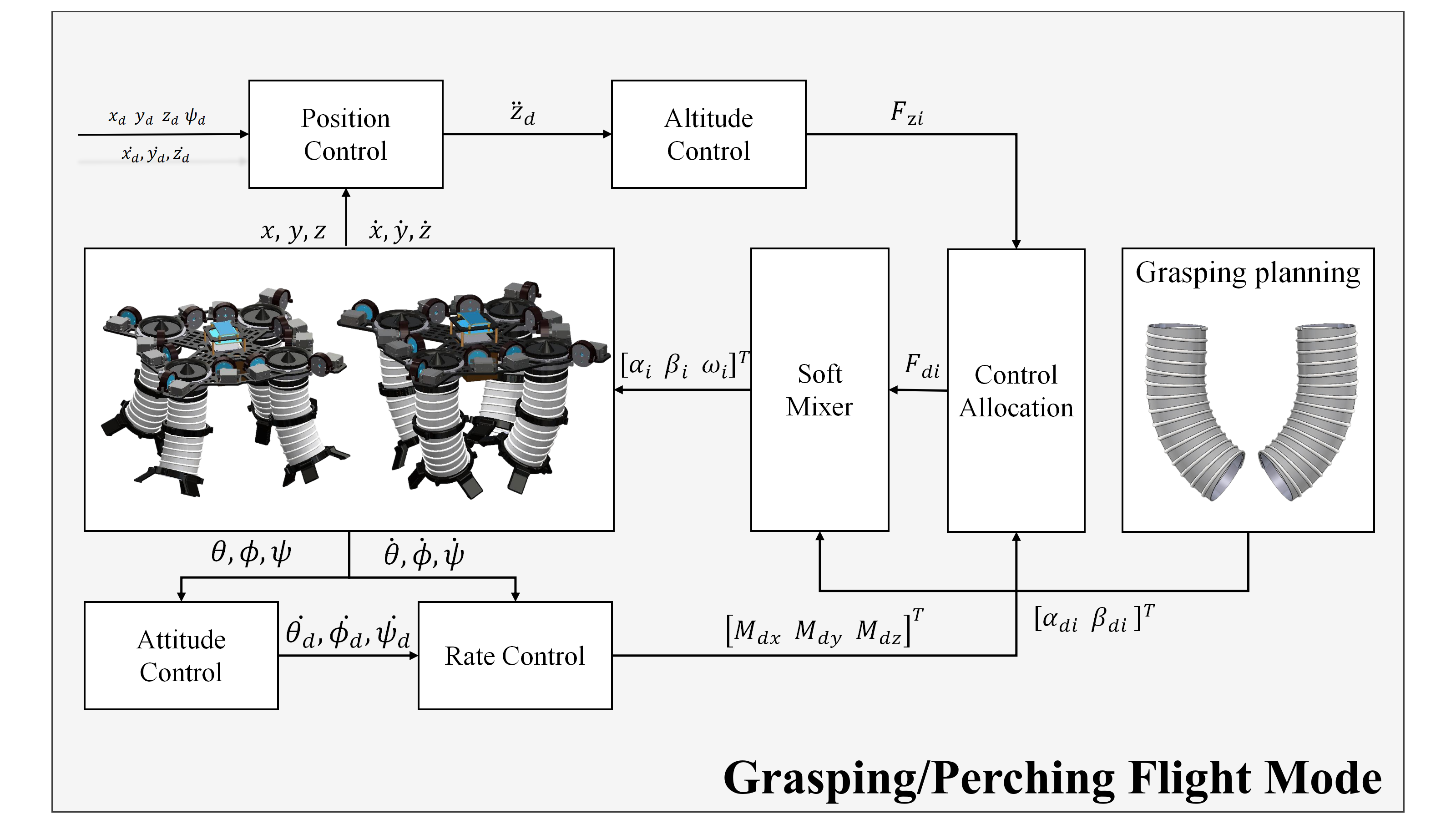}
\caption{The architecture of under-actuated flight Controller}
\label{fig_8}
\end{figure}

\textbf{Grasping planning:} In order to adapt to the shape of the object to be grasped, each SVPN must be extended in advance which ensures that the four SVPNs can fully envelop the target object. The SVPNs then bend towards the object converegently and fix their bending angle. The inclination angle generated by the grasp is determined by the grasping/perching planner, which provides the mixer with the desired angle of SVPN to bend the nozzles and complete the grasp.Once the grasping maneuver has been completed, Flexbee locks angles $\alpha$ and $\beta$. After this,it can only be controlled by adjusting the rotational speed $\omega$ to regulate vertical force $F_z$, as well as roll, yaw and pitch. In this mode, Flexbee operates as an under-actuated UAV, with dynamics and flight control that are analogous to a quadcopter. The controller of this mode will be experimentally validated in Chapter \uppercase\expandafter{\romannumeral5}.

\section{Experiments}
In this section, we will verify Flexbee's fully-actuated flight, under-actuated flight, and grasping/perching capabilities in an experimental environment, as well as its dynamics model and control strategy. The parameters of Flexbee as mentioned in the formula designed in this study are shown below:
$m=1.2 kg$, $g=9.8 m/s^2$, $I_{xx}=0.00913 kg \cdot m^2$, $I_{yy}=0.00918 kg \cdot m^2$, $I_{zz}=0.01245 kg \cdot m^2$,
$h=2.5 cm$, $s=12.0 cm$, $l=10.0 cm$.
Other performance parameters for Flexbee and SVPN are shown as below:
$\theta_{max}=30^{\circ}$, $\phi_{max}=30^{\circ}$, $\alpha_{max}=45^{\circ}$, $s_{max}=15 cm$.
More detailed experimental information can be found in the supplementary video materials.

\begin{figure}[!t]
\includegraphics[width=1\linewidth]{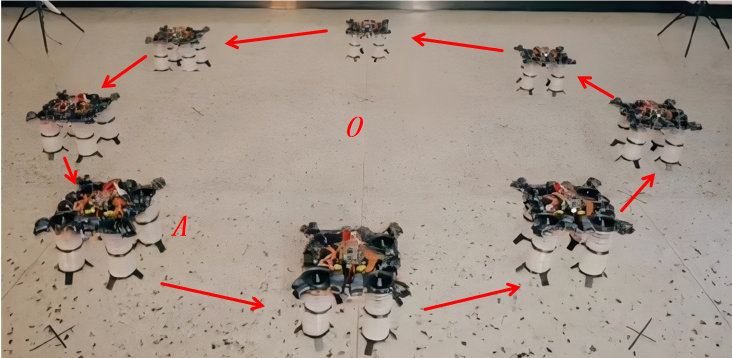}
\caption{The actual flight path of Flexbee}
\label{fig_9}

\includegraphics[width=1\linewidth]{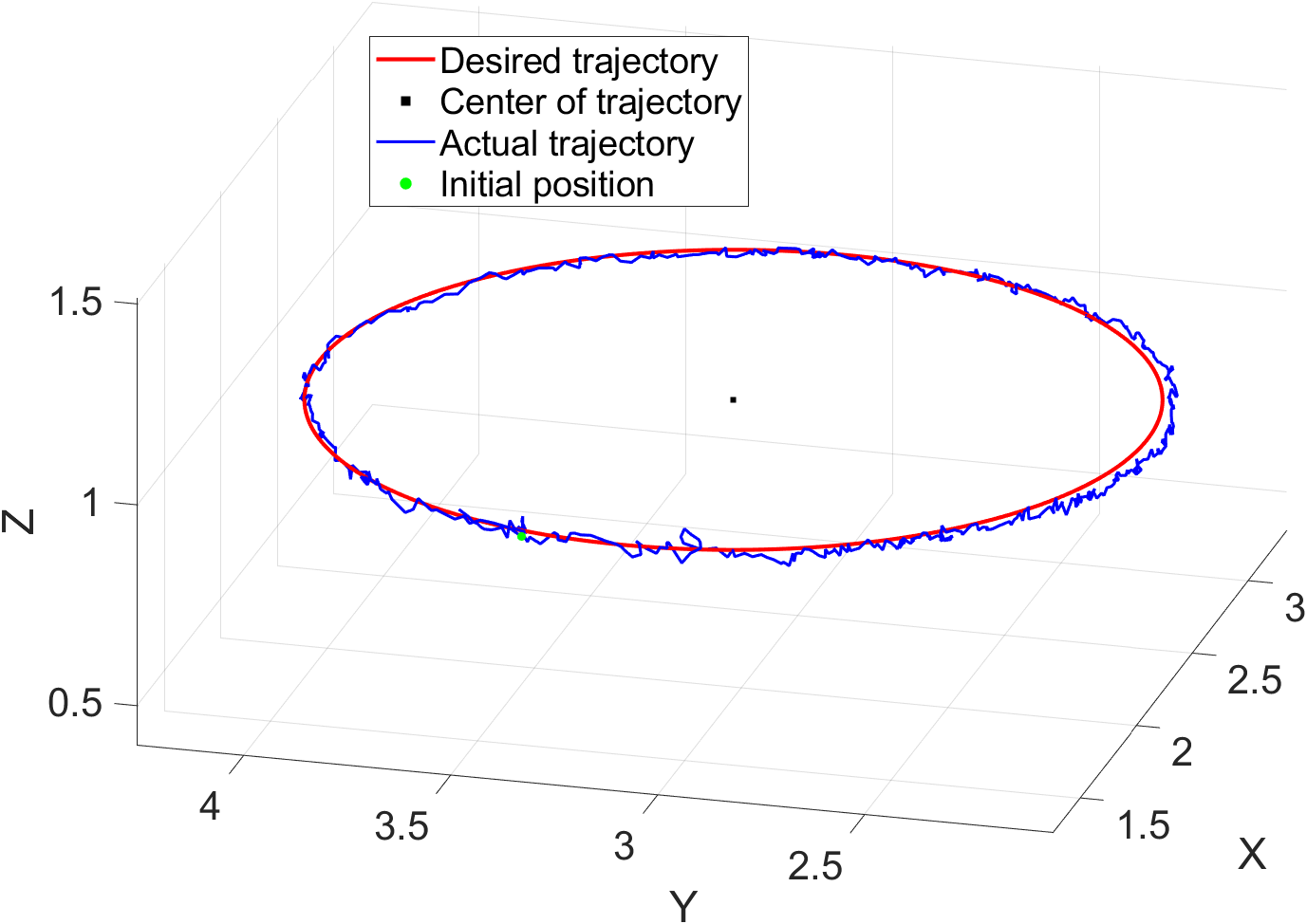}
\caption{Compares the actual trajectory recorded during the flight with the desired trajectory.}
\label{fig_10}

\includegraphics[width=1\linewidth]{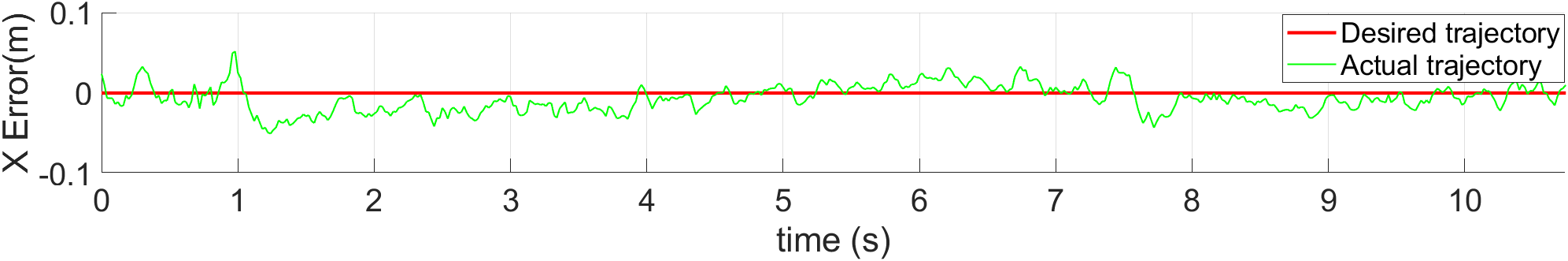}
\caption{Depicts the changes in positional error along the $X$ axes over time.}
\label{fig_11}

\includegraphics[width=1\linewidth]{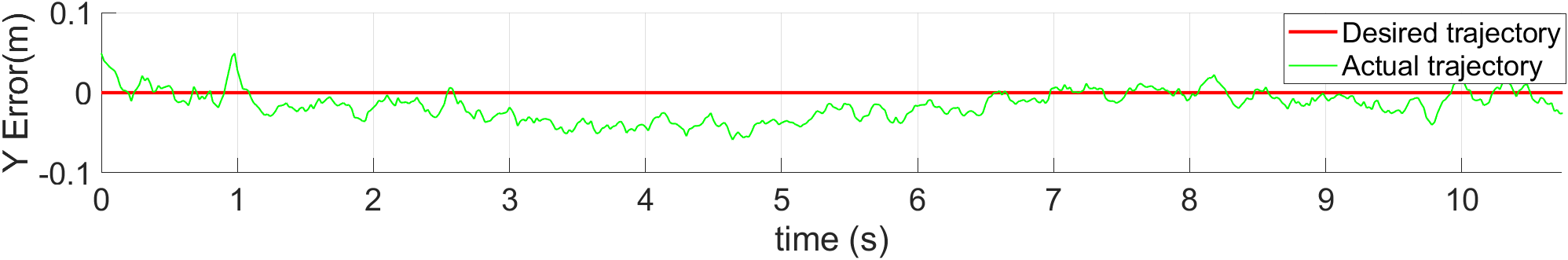}
\caption{Depicts the changes in positional error along the $Y$ axes over time.}
\label{fig_12}

\includegraphics[width=1\linewidth]{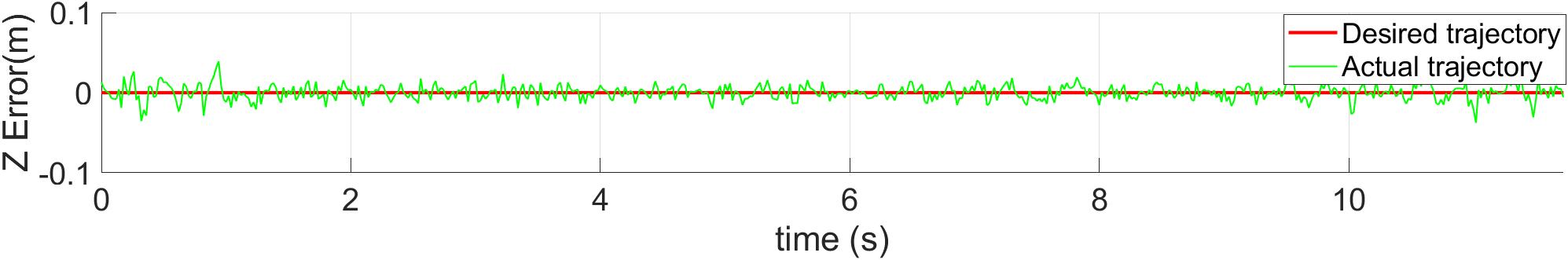}
\caption{Depicts the changes in positional error along the $Z$ axes over time.}
\label{fig_13}

\includegraphics[width=1\linewidth]{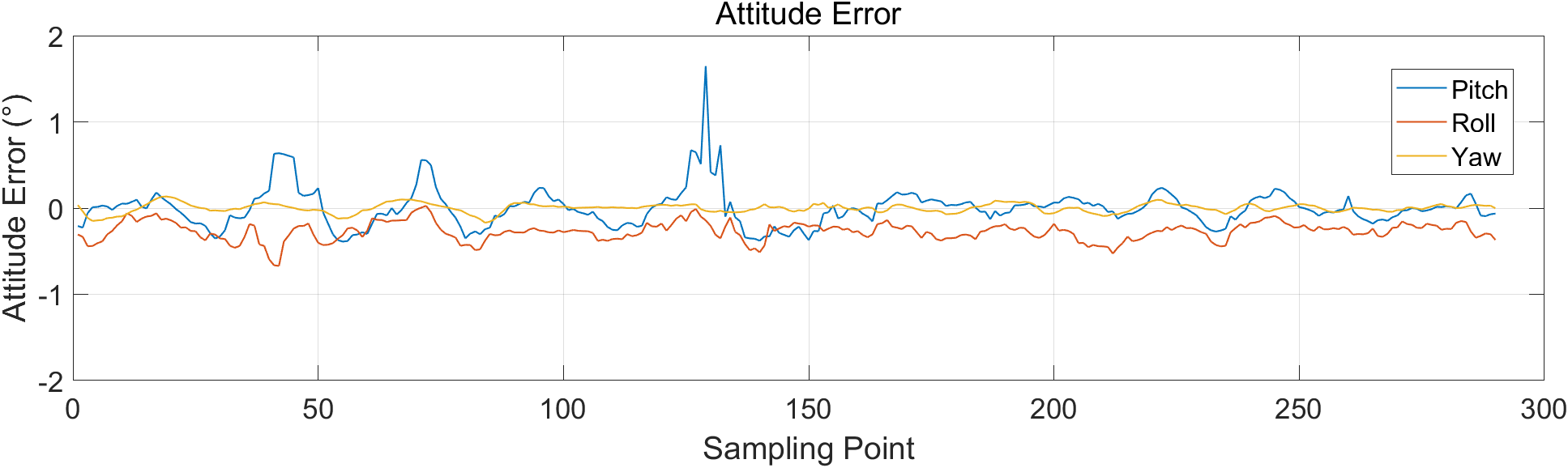}
\caption{$pitch$, $roll$ and $yaw$ change during the experiment.}
\label{fig_14}
\end{figure}

\subsection{Fully-actuated Flight Experiment}
In the first experiment, the control aircraft tracked a 6-D circular trajectory. With the lower left corner of the test site as the origin of the world coordinate system, the target trajectory is a horizontal circle with a radius of 1m, moving at 1m from the ground, with the centre of the trajectory circle at (2.3 2.8 1.0), and the control aircraft is always stable at the attitude angle, i.e., the desired roll, yaw, and pitch is 0. 

The experiment results are shown in Fig.\ref{fig_9}, Fig.\ref{fig_10}, Fig.\ref{fig_11}, Fig.\ref{fig_12}, Fig.\ref{fig_13}, and Fig.\ref{fig_14}. The experimental demonstration is shown in the video of the Supplementary Material. As shown in Fig.\ref{fig_9}, the actual flight path of Flexbee is presented.  Fig.\ref{fig_10} shows the position tracking comparison of the aircraft in 3D space, Fig.\ref{fig_11}, Fig.\ref{fig_12}, and Fig.\ref{fig_13},represents the co-ordinate change of the aircraft on the $X$-axis, $Y$-axis and $Z$-axis. Fig.\ref{fig_14} shows the actual roll, yaw, and pitch of the aircraft.

The experimental results demonstrate that the position and attitude of the aircraft can be controlled independently, thus confirming the decoupling of position and attitude dynamics in this fully-actuated vehicle. Furthermore, the aircraft control model is verified for fully-actuated flight mode.

\subsection{Under-actuated flight Experiment}
When switching to the grasping/perching flight mode, Flexbee fixes the bending angles of its four SVPNs. At this point, Flexbee operates in under-actuated flight mode, where its altitude and attitude can only be controlled by adjusting the rotational speed of the four propellers. To ensure attitude stability during under-actuated flight, the experimental setup transitions Flexbee from fully-actuated flight mode to under-actuated flight mode while maintaining a locked altitude. Flexbee sustains stable attitude in under-actuated flight mode and retains the capability for flight. At this stage, the control of it is similar to a quadcopter, which is not elaborated further here.

The results show that the aircraft can maintain attitude stability during the transition from fully-actuated to under-actuated flight mode, while retaining controllability in under-actuated mode.

\subsection{Grasping and Perching Experiment}
To verify the reliability and effectiveness of the grasping/perching mode, Flexbee underwent a series of experiments: the grasping experiment, the perching experiment, and the grasping-perching-flight experiment.
\begin{figure}[b]
\centering
\subfloat[]{\includegraphics[width=1.7in]{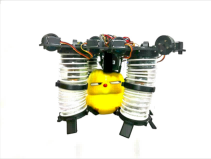}}%
\label{fig_first_case}
\subfloat[]{\includegraphics[width=1.7in]{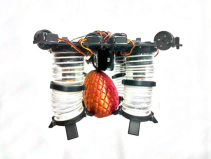}}%
\label{fig_second_case}
\subfloat[]{\includegraphics[width=1.7in]{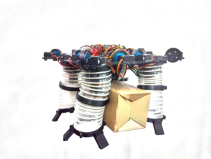}}%
\label{fig_third_case}
\subfloat[]{\includegraphics[width=1.7in]{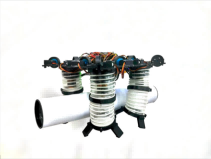}}%
\label{fig_forth_case}
\caption{Flexbee can grasp a variety of objects. (a) irregular soft objects  (b) hard spherical objects (c) objects larger than the inner diameter of the airframe. (d) objects smaller than the inner diameter of the airframe.}
\label{fig_15}
\end{figure}

In the grasping experiment, the adaptability of Flexbee's grasping mechanism was assessed by evaluating its ability to grasp various types of objects, including irregular soft objects, hard spherical objects, and objects larger or smaller than the inner diameter of the airframe. The experimental results are presented in Fig.\ref{fig_15}. The results demonstrate that Flexbee is capable of effectively enveloping and grasping objects of different shapes, materials, and sizes.

\begin{figure}
\centering
\subfloat[]{\includegraphics[width=2in]{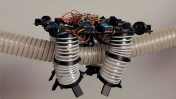}}%
\label{fig_first_case1}
\subfloat[]{\includegraphics[width=1.4in]{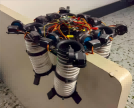}}%
\label{fig_second_case2}
\subfloat[]{\includegraphics[width=1.95in]{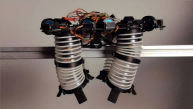}}%
\label{fig_third_case3}
\subfloat[]{\includegraphics[width=1.45in]{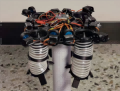}}%
\label{fig_forth_case4}
\caption{Flexbee can perform perching and landing in a variety of situations. (a) perching on a pipe (b) perching on a vertical flat plate, (c) perching on a pole (d) perching on a vertical projection}
\label{fig_16}
\end{figure}

In the perching performance experiments, the adaptability of Flexbee's perching ability was evaluated by testing it on various surfaces, including pipes, boards, poles, and cones. The experimental results are presented in Fig.\ref{fig_16}.

The results demonstrate that the aircraft exhibits strong environmental adaptability when perched on different surfaces. Flexbee will adapt to different environments by adjusting the extension or bending of SVPNs to enhance operational reliability.

\begin{figure}
\centering
\includegraphics[width=3.5in]{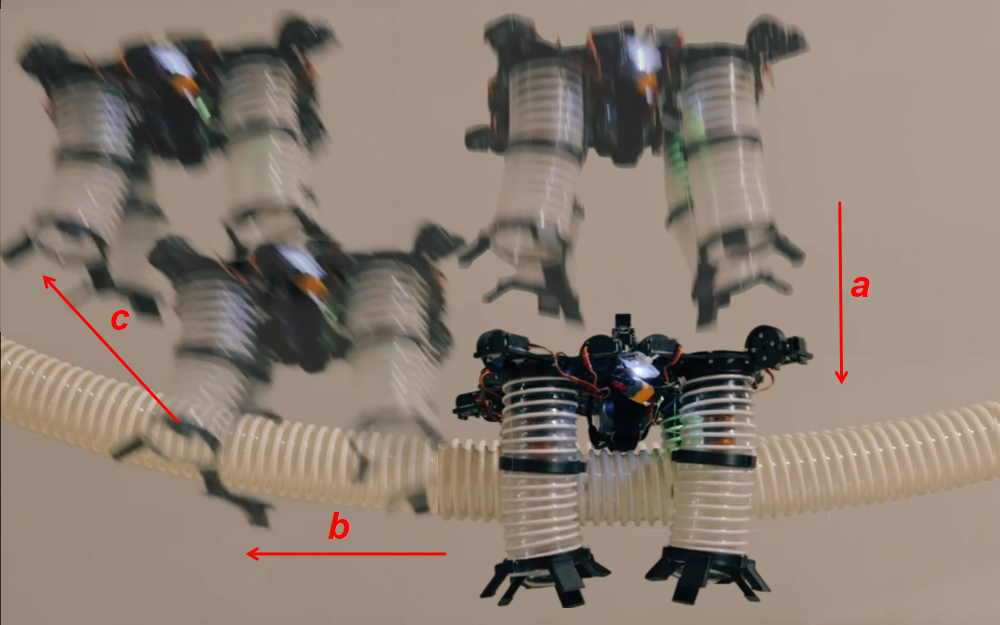}
\caption{The process of Flexbee's landing, perching and taking off.}
\label{fig_17}
\end{figure}

In the grasping-perching-flight experiment, Flexbee flew from the ground to the top of the pipe. It then extended its SVPNs radially, gradually landing on the pipe. Finally, it bent its SVPNs to achieve a stable perch. When required, it took off from the pipe by extending its SVPNs and flying away  The experimental results are presented in Fig.\ref{fig_17}:

Process $a$: Flexbee extended its SVPNs, landed on the pipeline, and achieved a grasping perching.
Process $b$: Flexbee extended its SVPNs with increasing the throttle, and took off from the pipeline.
Process $c$: Flexbee regained fully-actuated flight capability after completely detaching from the pipeline.

The results demonstrate that Flexbee can successfully execute a sequence of maneuvers, including fully-actuated flight, under-actuated flight for grasping and perching. This confirms the effectiveness of the aircraft's dynamics model and control strategy across all operational modes.

\section{Conclusion}
This paper explores the mechanical design, modeling, and control strategy of the grasping and perching UAV, Flexbee, based on soft vector propulsion nozzle. The SVPN that combines motion control capabilities for generating vector force with structurally adaptive grasping and perching capabilities is proposed as a first attempt, significantly enhancing the UAV's motion control performance, structural reusability, and environmental adaptability. The dynamics modeling and controller design based on this structure have been validated through Flexbee's fully-actuated flight mode and grasping/perching flight mode experiments, achieving precise and flexible control of flight, grasping and perching. The experimental results highlight the potential of Flexbee and demonstrate the effectiveness of the designed controller. Future work will focus on developing the flight trajectory planning for the entire grasping and perching process, as well as a quadrupedal gait mode for Flexbee.

\bibliographystyle{IEEEtran}
\bibliography{ref}

\begin{thebibliography}{10}
\providecommand{\url}[1]{#1}
\csname url@samestyle\endcsname
\providecommand{\newblock}{\relax}
\providecommand{\bibinfo}[2]{#2}
\providecommand{\BIBentrySTDinterwordspacing}{\spaceskip=0pt\relax}
\providecommand{\BIBentryALTinterwordstretchfactor}{4}
\providecommand{\BIBentryALTinterwordspacing}{\spaceskip=\fontdimen2\font plus
\BIBentryALTinterwordstretchfactor\fontdimen3\font minus
  \fontdimen4\font\relax}
\providecommand{\BIBforeignlanguage}[2]{{%
\expandafter\ifx\csname l@#1\endcsname\relax
\typeout{** WARNING: IEEEtran.bst: No hyphenation pattern has been}%
\typeout{** loaded for the language `#1'. Using the pattern for}%
\typeout{** the default language instead.}%
\else
\language=\csname l@#1\endcsname
\fi
#2}}
\providecommand{\BIBdecl}{\relax}
\BIBdecl

\bibitem{ref1}
J.~Yang, Y.~Zhu, L.~Zhang, Y.~Dong, and Y.~Ding, ``Sytab: A class of
  smooth-transition hybrid terrestrial/aerial bicopters,'' \emph{IEEE Robotics
  and Automation Letters}, vol.~7, no.~4, pp. 9199--9206, 2022.

\bibitem{ref2}
P.~Rudol and P.~Doherty, ``Human body detection and geolocalization for uav
  search and rescue missions using color and thermal imagery,'' in \emph{2008
  IEEE Aerospace Conference}, 2008, pp. 1--8.

\bibitem{ref3}
J.~Chen, T.~Liu, and S.~Shen, ``Tracking a moving target in cluttered
  environments using a quadrotor,'' in \emph{2016 IEEE/RSJ International
  Conference on Intelligent Robots and Systems (IROS)}, 2016, pp. 446--453.

\bibitem{ref4}
J.~Casper and R.~Murphy, ``Human-robot interactions during the robot-assisted
  urban search and rescue response at the world trade center,'' \emph{IEEE
  Transactions on Systems, Man, and Cybernetics, Part B (Cybernetics)},
  vol.~33, no.~3, pp. 367--385, 2003.

\bibitem{ref5}
S.~Minaeian, J.~Liu, and Y.-J. Son, ``Vision-based target detection and
  localization via a team of cooperative uav and ugvs,'' \emph{IEEE
  Transactions on Systems, Man, and Cybernetics: Systems}, vol.~46, no.~7, pp.
  1005--1016, 2016.

\bibitem{ref6}
A.~Patrik, G.~Utama, A.~A.~S. Gunawan, A.~Chowanda, J.~S. Suroso,
  R.~Shofiyanti, and W.~Budiharto, ``Gnss-based navigation systems of
  autonomous drone for delivering items,'' \emph{Journal of Big Data}, vol.~6,
  no.~1, p.~53, 2019.

\bibitem{ref7}
K.~Appeaning~Addo, P.-N. Jayson-Quashigah, S.~N.~A. Codjoe, and F.~Martey,
  ``Drone as a tool for coastal flood monitoring in the volta delta, ghana,''
  \emph{Geoenvironmental Disasters}, vol.~5, no.~1, pp. 1--13, 2018.

\bibitem{ref8}
\BIBentryALTinterwordspacing
C.~Gomez and H.~Purdie, ``Uav-based photogrammetry and geocomputing for hazards
  and disaster risk monitoring – a review,'' \emph{Geoenvironmental
  Disasters}, vol.~3, no.~1, p.~23, 2016. [Online]. Available:
  \url{https://doi.org/10.1186/s40677-016-0060-y}
\BIBentrySTDinterwordspacing

\bibitem{ref9}
W.~R. Roderick, M.~R. Cutkosky, and D.~Lentink, ``Touchdown to take-off: at the
  interface of flight and surface locomotion,'' \emph{Interface focus}, vol.~7,
  no.~1, p. 20160094, 2017.

\bibitem{ref10}
J.~Meng, J.~Buzzatto, Y.~Liu, and M.~Liarokapis, ``On aerial robots with
  grasping and perching capabilities: A comprehensive review,'' \emph{Frontiers
  in Robotics and AI}, vol.~8, p. 739173, 2022.

\bibitem{ref11}
H.~W. Wopereis, T.~Van Der~Molen, T.~H. Post, S.~Stramigioli, and M.~Fumagalli,
  ``Mechanism for perching on smooth surfaces using aerial impacts,'' in
  \emph{2016 IEEE international symposium on safety, security, and rescue
  robotics (SSRR)}.\hskip 1em plus 0.5em minus 0.4em\relax IEEE, 2016, pp.
  154--159.

\bibitem{ref12}
P.~Liu, S.~P. Sane, J.-M. Mongeau, J.~Zhao, and B.~Cheng, ``Flies land upside
  down on a ceiling using rapid visually mediated rotational maneuvers,''
  \emph{Science advances}, vol.~5, no.~10, p. eaax1877, 2019.

\bibitem{ref13}
S.~Liu, W.~Dong, Z.~Ma, and X.~Sheng, ``Adaptive aerial grasping and perching
  with dual elasticity combined suction cup,'' \emph{IEEE Robotics and
  Automation Letters}, vol.~5, no.~3, pp. 4766--4773, 2020.

\bibitem{ref14}
H.-N. Nguyen, R.~Siddall, B.~Stephens, A.~Navarro-Rubio, and M.~Kova{\v{c}},
  ``A passively adaptive microspine grapple for robust, controllable
  perching,'' in \emph{2019 2nd IEEE international conference on soft robotics
  (RoboSoft)}.\hskip 1em plus 0.5em minus 0.4em\relax IEEE, 2019, pp. 80--87.

\bibitem{ref15}
A.~Lussier~Desbiens and M.~R. Cutkosky, ``Landing and perching on vertical
  surfaces with microspines for small unmanned air vehicles,'' \emph{Journal of
  Intelligent and Robotic Systems}, vol.~57, no.~1, pp. 313--327, 2010.

\bibitem{ref16}
T.~Kominami and K.~Shimonomura, ``Versatile perching using a passive mechanism
  with under actuated fingers for multirotor uav,'' \emph{IEEE Robotics and
  Automation Letters}, 2024.

\bibitem{ref17}
C.~E. Doyle, J.~J. Bird, T.~A. Isom, J.~C. Kallman, D.~F. Bareiss, D.~J.
  Dunlop, R.~J. King, J.~J. Abbott, and M.~A. Minor, ``An avian-inspired
  passive mechanism for quadrotor perching,'' \emph{IEEE/ASME Transactions On
  Mechatronics}, vol.~18, no.~2, pp. 506--517, 2012.

\bibitem{ref18}
M.~L. Burroughs, K.~Beauwen~Freckleton, J.~J. Abbott, and M.~A. Minor, ``A
  sarrus-based passive mechanism for rotorcraft perching,'' \emph{Journal of
  Mechanisms and Robotics}, vol.~8, no.~1, p. 011010, 2016.

\bibitem{ref19}
D.~J. Dunlop and M.~A. Minor, ``Modeling and simulation of perching with a
  quadrotor aerial robot with passive bio-inspired legs and feet,'' \emph{ASME
  Letters in Dynamic Systems and Control}, vol.~1, no.~2, p. 021005, 2021.

\bibitem{ref20}
W.~Chi, K.~Low, K.~H. Hoon, and J.~Tang, ``An optimized perching mechanism for
  autonomous perching with a quadrotor,'' in \emph{2014 IEEE international
  conference on robotics and automation (ICRA)}.\hskip 1em plus 0.5em minus
  0.4em\relax IEEE, 2014, pp. 3109--3115.

\bibitem{ref21}
K.~M. Popek, M.~S. Johannes, K.~C. Wolfe, R.~A. Hegeman, J.~M. Hatch, J.~L.
  Moore, K.~D. Katyal, B.~Y. Yeh, and R.~J. Bamberger, ``Autonomous grasping
  robotic aerial system for perching (agrasp),'' in \emph{2018 IEEE/RSJ
  international conference on intelligent robots and systems (IROS)}.\hskip 1em
  plus 0.5em minus 0.4em\relax IEEE, 2018, pp. 1--9.

\bibitem{ref22}
U.~A. Fiaz, M.~Abdelkader, and J.~S. Shamma, ``An intelligent gripper design
  for autonomous aerial transport with passive magnetic grasping and
  dual-impulsive release,'' in \emph{2018 IEEE/ASME International Conference on
  Advanced Intelligent Mechatronics (AIM)}.\hskip 1em plus 0.5em minus
  0.4em\relax IEEE, 2018, pp. 1027--1032.

\bibitem{ref23}
P.~M. Nadan, T.~M. Anthony, D.~M. Michael, J.~B. Pflueger, M.~S. Sethi, K.~N.
  Shimazu, M.~Tieu, and C.~L. Lee, ``A bird-inspired perching landing gear
  system,'' \emph{Journal of Mechanisms and Robotics}, vol.~11, no.~6, 2019.

\bibitem{ref24}
A.~McLaren, Z.~Fitzgerald, G.~Gao, and M.~Liarokapis, ``A passive closing,
  tendon driven, adaptive robot hand for ultra-fast, aerial grasping and
  perching,'' in \emph{2019 IEEE/RSJ International Conference on Intelligent
  Robots and Systems (IROS)}.\hskip 1em plus 0.5em minus 0.4em\relax IEEE,
  2019, pp. 5602--5607.

\bibitem{ref25}
C.~C. Kessens, J.~Thomas, J.~P. Desai, and V.~Kumar, ``Versatile aerial
  grasping using self-sealing suction,'' in \emph{2016 IEEE international
  conference on robotics and automation (ICRA)}.\hskip 1em plus 0.5em minus
  0.4em\relax IEEE, 2016, pp. 3249--3254.

\bibitem{ref26}
M.~W. Hannan and I.~D. Walker, ``Kinematics and the implementation of an
  elephant's trunk manipulator and other continuum style robots,''
  \emph{Journal of robotic systems}, vol.~20, no.~2, pp. 45--63, 2003.

\bibitem{ref27}
C.~Armanini, F.~Boyer, A.~T. Mathew, C.~Duriez, and F.~Renda, ``Soft robots
  modeling: A structured overview,'' \emph{IEEE Transactions on Robotics},
  vol.~39, no.~3, pp. 1728--1748, 2023.

\end{thebibliography}

\vspace{1pt}
\begin{IEEEbiography}[{\includegraphics[width=1in,height=1.25in,clip,keepaspectratio]{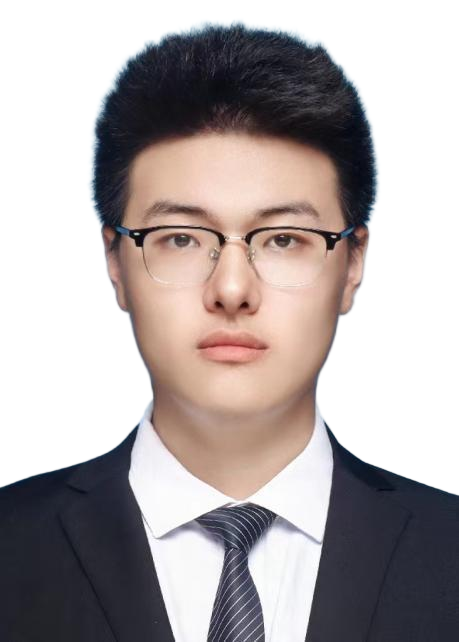}}]{Yue Wang}
   received the B.S. degree in communications engineering from the Ocean University of China, Qingdao, China, in 2023. He is currently working toward the Ph.D. degree in aerospace science and technology with the Harbin Institute of Technology, Harbin, China. 

   His research interests include the design, modeling and control of multimodal aircraft, as well as soft and underwater robotics.
 \end{IEEEbiography}
 
 \vspace{1pt}
 \begin{IEEEbiography}[{\includegraphics[width=1in,height=1.25in,clip,keepaspectratio]{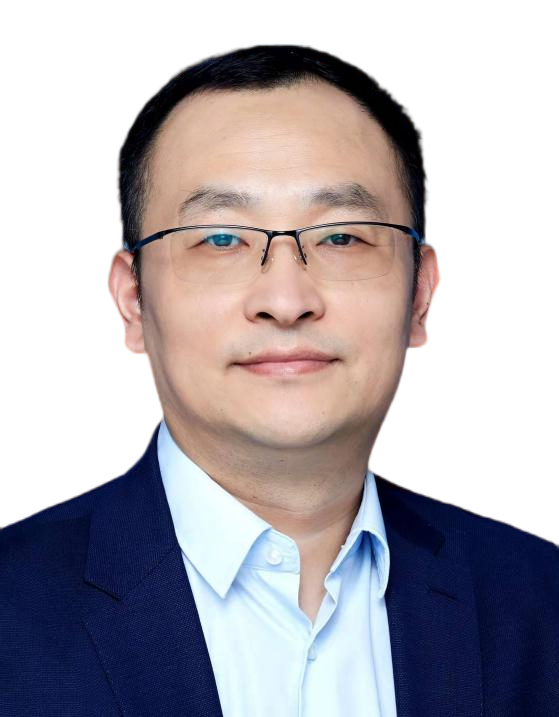}}]{Lixian Zhang}
 	(Fellow, IEEE) received the Ph.D. degree in control science and engineering from the Harbin Institute of Technology, Harbin, China, in 2006. 
 	
 	Since 2009, he has been with the Harbin Institute of Technology, Harbin, China, where he is currently a professor. His research interests include nondeterministic switched systems, model predictive control and their applications in multimodal unmanned systems.
 \end{IEEEbiography}
  \vspace{1pt}
 \begin{IEEEbiography}
 [{\includegraphics[width=1in,height=1.25in,clip,keepaspectratio]{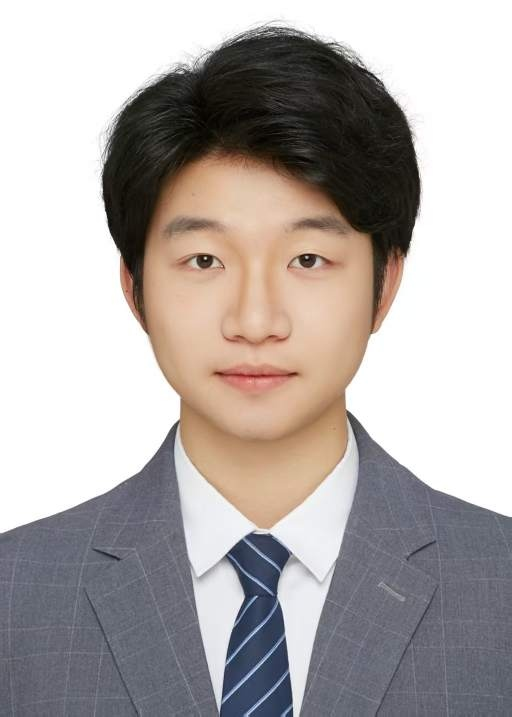}}]{Yimin Zhu}
   received the B.S. degree in automation from the Harbin Engineering University, Harbin, China, in 2020. He is currently working toward the Ph.D. degree in aerospace science and technology with the Harbin Institute of Technology, Harbin, China. 

   His research interests include robotics, discrete event systems, and model predictive control.
 \end{IEEEbiography}
\vspace{1pt}
\begin{IEEEbiography}[{\includegraphics[width=1in,height=1.25in,clip,keepaspectratio]{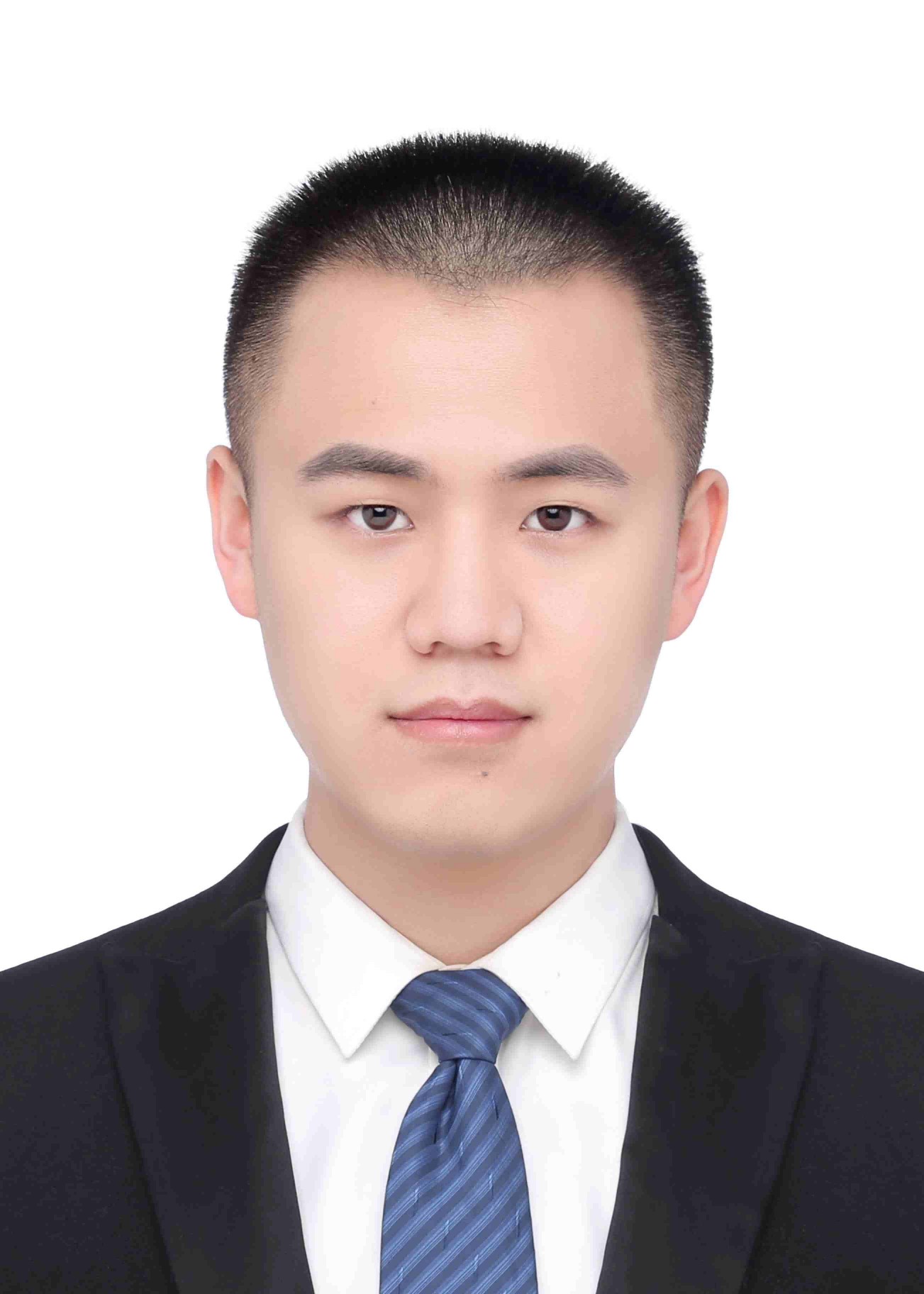}}]{Yangguang Liu}
   received the B.S. degree in electronic and information engineering from the Harbin Engineering University, Harbin, China, in 2023. He is currently working toward the Ph.D. degree in control Science and Engineering with the Harbin Institute of Technology, Harbin, China. 

   His research interests include robotics, design and control of hybrid terrestrial-aerial aircraft, and unmanned underwater vehicle design.
 \end{IEEEbiography}
 \vspace{1pt} \begin{IEEEbiography}[{\includegraphics[width=1in,height=1.25in,clip,keepaspectratio]{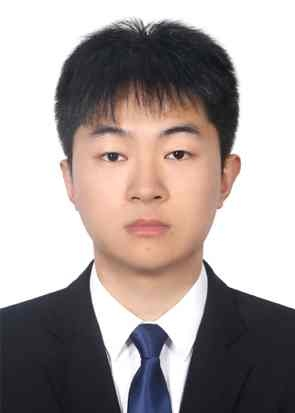}}]{Xuwei Yang}
Xuwei Yang received the B.E. degree in School of Astronautics from Harbin Institute Technology in 2025.He is currently pursuing a Doctor's degree in Control Engineering at Harbin Institute of Technology,Harbin,China.His research interests include control theory and robotics.
 \end{IEEEbiography}
\end{document}